\PassOptionsToPackage{unicode}{hyperref}
\PassOptionsToPackage{hyphens}{url}
\documentclass[
]{article}
\usepackage{amsmath,amssymb}
\usepackage{iftex}
\ifPDFTeX
  \usepackage[T1]{fontenc}
  \usepackage[utf8]{inputenc}
  \usepackage{textcomp} 
\else 
  \usepackage{unicode-math} 
  \defaultfontfeatures{Scale=MatchLowercase}
  \defaultfontfeatures[\rmfamily]{Ligatures=TeX,Scale=1}
\fi
\usepackage{lmodern}
\ifPDFTeX\else
\fi
\IfFileExists{upquote.sty}{\usepackage{upquote}}{}
\IfFileExists{microtype.sty}{
  \usepackage[]{microtype}
  \UseMicrotypeSet[protrusion]{basicmath} 
}{}
\makeatletter
\@ifundefined{KOMAClassName}{
  \IfFileExists{parskip.sty}{%
    \usepackage{parskip}
  }{
    \setlength{\parindent}{0pt}
    \setlength{\parskip}{6pt plus 2pt minus 1pt}}
}{
  \KOMAoptions{parskip=half}}
\makeatother
\usepackage{xcolor}
\usepackage[margin=1in]{geometry}
\usepackage{color}
\usepackage{fancyvrb}

\DefineVerbatimEnvironment{Highlighting}{Verbatim}{commandchars=\\\{\}}
\usepackage{framed}
\definecolor{shadecolor}{RGB}{248,248,248}
\newenvironment{Shaded}{\begin{snugshade}}{\end{snugshade}}

\newcommand{\AttributeTok}[1]{\textcolor[rgb]{0.13,0.29,0.53}{#1}}

\newcommand{\CommentTok}[1]{\textcolor[rgb]{0.56,0.35,0.01}{\textit{#1}}}

\newcommand{\ConstantTok}[1]{\textcolor[rgb]{0.56,0.35,0.01}{#1}}

\newcommand{\DecValTok}[1]{\textcolor[rgb]{0.00,0.00,0.81}{#1}}

\newcommand{\FloatTok}[1]{\textcolor[rgb]{0.00,0.00,0.81}{#1}}
\newcommand{\FunctionTok}[1]{\textcolor[rgb]{0.13,0.29,0.53}{\textbf{#1}}}

\newcommand{\NormalTok}[1]{#1}

\newcommand{\OtherTok}[1]{\textcolor[rgb]{0.56,0.35,0.01}{#1}}

\newcommand{\SpecialCharTok}[1]{\textcolor[rgb]{0.81,0.36,0.00}{\textbf{#1}}}

\newcommand{\StringTok}[1]{\textcolor[rgb]{0.31,0.60,0.02}{#1}}

\usepackage{longtable,booktabs,array}
\usepackage{calc} 
\usepackage{etoolbox}
\makeatletter
\patchcmd\longtable{\par}{\if@noskipsec\mbox{}\fi\par}{}{}
\makeatother
\IfFileExists{footnotehyper.sty}{\usepackage{footnotehyper}}{\usepackage{footnote}}
\makesavenoteenv{longtable}
\usepackage{graphicx}
\makeatletter
\def\maxwidth{\ifdim\Gin@nat@width>\linewidth\linewidth\else\Gin@nat@width\fi}
\def\maxheight{\ifdim\Gin@nat@height>\textheight\textheight\else\Gin@nat@height\fi}
\makeatother
\setkeys{Gin}{width=\maxwidth,height=\maxheight,keepaspectratio}
\makeatletter
\def\fps@figure{htbp}
\makeatother
\providecommand{\tightlist}{%
  \setlength{\itemsep}{0pt}\setlength{\parskip}{0pt}}
\ifLuaTeX
  \usepackage{selnolig}  
\fi
\IfFileExists{bookmark.sty}{\usepackage{bookmark}}{\usepackage{hyperref}}
\IfFileExists{xurl.sty}{\usepackage{xurl}}{} 
\hypersetup{
  pdftitle={Assessing Keyness using Permutation Tests},
  pdfauthor={Thoralf Mildenberger},
  hidelinks,
  pdfcreator={LaTeX via pandoc}}

\title{Assessing Keyness using Permutation Tests}
\author{Thoralf Mildenberger\footnote{Institute of Data Analysis and
  Process Design, ZHAW Zurich University of Applied Sciences,
  \href{mailto:mild@zhaw.ch}{\nolinkurl{mild@zhaw.ch}}}}
\date{2023-08-25}

\begin{document}
\maketitle
\begin{abstract}
We propose a resampling-based approach for assessing keyness in corpus
linguistics based on suggestions by Gries (2006, 2022). Traditional
approaches based on hypothesis tests (e.g.~Likelihood Ratio) model the
copora as independent identically distributed samples of tokens. This
model does not account for the often observed uneven distribution of
occurences of a word across a corpus. When occurences of a word are
concentrated in few documents, large values of LLR and similar scores
are in fact much more likely than accounted for by the token-by-token
sampling model, leading to false positives. We replace the
token-by-token sampling model by a model where corpora are samples of
documents rather than tokens, which is much closer to the way corpora
are actually assembled. We then use a permutation approach to
approximate the distribution of a given keyness score under the null
hypothesis of equal frequencies and obtain p-values for assessing
significance. We do not need any assumption on how the tokens are
organized within or across documents, and the approach works with
basically \emph{any} keyness score. Hence, appart from obtaining more
accurate p-values for scores like LLR, we can also assess significance
for e.g.~the logratio which has been proposed as a measure of effect
size. An efficient implementation of the proposed approach is provided
in the \texttt{R} package \texttt{keyperm} available from github.
\end{abstract}

\hypertarget{introduction}{%
\section{Introduction}\label{introduction}}

In this paper, we consider the \emph{keyness problem}, namely, assessing
whether some words appear significantly more often in one corpus \(A\)
than in another corpus \(B\). Most existing approaches are based on
\emph{statistical hypothesis tests}, i.e.~differences in frequencies of
a word between two corpora \(A\) and \(B\) are judged to be significant
if they are so large that they would be very unlikely under a
\emph{random sampling model}. Examples of these approaches are the
well-known \emph{Log-Likelihood-Ratio-Test}, \emph{\(\chi^2\)-tests} and
\emph{Fisher's Exact Test}. Other measures like the so-called
\emph{Log-Ratio} are also sometimes used, but these do not directly take
into account random variation in the sampling process and hence tend to
give very high scores to differences in very rare words which might well
be due to chance.

Approaches based on statistical hypothesis tests are necessarily tied to
some assumption of randomness. All methods currently used -- as far as
they are based on hypothesis testing -- are justified under the
assumption that the corpora \(A\) and \(B\) are samples of larger
populations, say \(Pop_A\) and \(Pop_B\). These are typically not larger
corpora of actually existing texts from which a random selection was
taken but some form of abstract infinite populations like ``all texts
that could have been produced by some author'', ``all texts that could
have been published in a given newspaper in a given year'', ``the
discourse on\ldots{}'', ``actual use of language in certain media'' etc.
This is not a special feature of linguistics but very common in other
applications of statistics, as we typically view a sample of
e.g.~patients with a certain type of disease not as a subset of all
people who currently have the desease but also representative of
patients that will develop the illness in the future, some of whom might
not even be born yet. Hence, assuming the corpus is a random sample from
some larger (abstract) population is not a problem per se.

The problem is rather the specific random sampling mechanism that is
assumed. Here, we follow an argument also put forward in Gries (2006,
2022); see also Evert (2006) for a discussion of randomness and
different sampling models. While assuming the corpora to be random
samples of \emph{texts} from larger populations of (potentially
fictitious) texts seems reasonable in many cases, the assumption
actually underlying the hypothesis tests as currently used is that the
corpora are samples of \emph{tokens}, i.e.~the corpora are modeled as
constructed by \emph{independently} drawing single tokens, one by one,
from the two populations. The reason is of course that the distributions
of the relevant test statistics are known at least approximately under
this assumption (\(\chi^2_1\) for \(LLR\)- and \(\chi^2\)-tests,
hypergeometric for Fisher's).

Apart from the fact that the smallest unit added to a corpus at a time
is always a whole text (perhaps a very short one like in sentence
corpora) this is unrealistic in at least two ways:

\begin{itemize}
\tightlist
\item
  Within a text, grammatical constraints limit which words can follow
  other words. The assumption of independence hence allows for
  ungrammatical as well as nonsensical texts.
\item
  Words are generally not evenly distributed in corpora, i.e.~they may
  appear several times in some texts and not at all in others.
\end{itemize}

While the first issue has also been used to criticize the use of
hypothesis testing in this context (Killgariff 2005), the second issue
seems to be more severe. The phenomenon itself is well known, and
measures have been developed to measure uneven distribution (see eg.
Gries (2008)). In addition, other procedures like topic modelling that
are routinely performed on the same corpora start from the assumption
that the distribution is uneven. Yet, the implications for keyness
analyses are routinely ignored.

It might not be obvious why sampling in units larger than single tokens
is a problem for keyness analyses at all. Word frequencies are measured
by counting occurences across the corpus and dividing by the total
number of tokens. This is also directly reflected in all currently used
measures of keyness. While the measures are rightly interpreted with
respect to tokens, as also pointed out by Gries (2006, 2022), the
sampling distributions of the measures are much wider when taking into
account that we sample \emph{texts}, not single \emph{tokens} at a time.
We show some empirical examples later, but the idea can be seen as
follows: Think of some word whose occurences are very unevenly spread,
i.e.~when we add a new text to the corpus, the occurence will be either
much less or much more than ``on average'', i.e.~the relative frequency
in the new text will greatly deviate from the relative frequency across
the whole population in either direction. On the other hand, if instead
we independently added a corresponding number of tokens independently
drawn, each of which has a probability of being equal to the word under
consideration equal to the relative frequency in the population, the
relative frequency of occurences will be much closer to the overall
average. Hence, the ratio of occurences to total size in tokens changes
much more when sampling larger units than it does when sampling single
tokens. However, the latter model is used for judging how extreme a
given score is, while more extreme values are actually much more likely
just by chance under the much more realistic model of sampling
text-by-text.

The increased variability due to not sampling independently
token-by-token cannot be amended by just setting stricter thresholds, as
the exact impact is generally unknown beforehand, may vary from corpus
to corpus and, more severely, between words within the same corpus,
depending on how unevenly distributed they are. Hence, two words might
have the same \(LLR\) score, but due to differences in dispersion one
might well be much more extreme than what would be expected by chance
while the other may not.

In the following, we propose an approach based on \emph{permutation
tests}, that allows us to simulate (to arbitrary precision) the
distribution of any conventionally used keyness score as well as new
ones under the null hypothesis of equal frequencies in the two
populations (technically speaking, we are actually testing a somewhat
stonger hypothesis). This means we are not tied to measures for which
the distribution is ``known'' (under a assumptions known to be blatantly
wrong), but we can use a measure that is also more interpretable. The
downside is that we cannot rely on standard critical values for judging
significance, as the distributions will be different for each word in
each corpus -- meaning we have to be willing to utilize more
computational resources. Permutation tests (see also Gries 2006) are
computationally very similar to the \emph{bootstrap} approach advocated
in Gries (2022), but are conceptually different and are based on
different assumptions.

The rest of the paper is organized as follows: In Section 2, we review
some of the theory behind hypothesis testing for keyness analysis and
explain the differences between the widely used sampling-token-by-token
model and our sampling-text-by-text model in more detail. In Section 3,
we point out some of the undesired effects of using current approaches
by means of a numerical example. In Section 4, we give some further
remarks on practical implementation.

The procedure introduced in this article is available in an R-package
\texttt{keyperm} that can be installed from github\footnote{The current
  development version can be installed by
  \texttt{remotes::install\_github("thmild/keyperm")}}. Submission to
CRAN is planned. In keyness analysis, we need high precision in the
p-values in order to compare them, hence we need many simulation runs.
To make this approach efficient, the core resampling part of the
procedure has been implemented in \texttt{C++} using the \texttt{Rcpp}
interface (Eddelbuettel et al.~2023).

\hypertarget{significance-testing-for-keyword-analysis}{%
\section{Significance testing for keyword
analysis}\label{significance-testing-for-keyword-analysis}}

We consider the following problem: We have two Corpora \(A\) and \(B\)
which are regarded as \emph{samples} from larger populations \(Pop_A\)
and \(Pop_B\) of (potential) texts. \emph{Keywords} are words which
appear more fruequently in \(Pop_A\) than \(Pop_B\) (sometimes we also
want those that are more frequent in \(Pop_B\)). Often, \emph{B} will be
a \emph{reference corpus}.

The general approach is a three-step procedure:

\begin{enumerate}
\def\labelenumi{\arabic{enumi}.}
\tightlist
\item
  Calculate a \emph{keyness score} for each word that appears at least
  once in one of \(A\) or \(B\) (usually directly equivalent to a
  \emph{p-value})
\item
  Filter the word list by setting a threshold on the scores (preferably
  taking into account multiple testing issues) and perhaps additional
  criteria
\item
  Order the words that pass the filter
\end{enumerate}

Steps 2 and 3 are often mixed, i.e.~the score is used for ordering the
list and words are excluded based on a cut-off for the same score. This
is not really appropriate as criteria like \(LLR\) measure
\emph{significance}, i.e.~strength of evidence against the null
hypothesis of equal frequencies in \(Pop_A\) and \(Pop_B\). As such,
they do not directly measure the \emph{strength of the effect} and are
therefore useful for filtering but not for ranking. Effect size measures
like the \emph{log-ratio} on the other hand do not take into account
random variation and tend to be extreme for very small numbers
occurences, making them unsuitable for filtering but useful for ranking.
Hence, it generally makes sense to \emph{filter by significance} and
\emph{order by effect size} (supported by \texttt{CQPweb} and recent
versions of \texttt{AntConc}).

We will almost exclusively focus on step 1 in this article, as the
currently used tests are not even valid for testing for a difference in
frequencies for a single word. Hence, it is most important to get the
tests (and associated \(p\)-values) right before considering problems of
combining the results (which involve issues of multiple testing).

\hypertarget{tests-of-significance-in-contingency-tables}{%
\subsection{Tests of significance in contingency
tables}\label{tests-of-significance-in-contingency-tables}}

For a potential keyword, consider the following contingency table:

\begin{longtable}[]{@{}lccc@{}}
\toprule\noalign{}
& freq. of \emph{word} & freq. of all other words & total \\
\midrule\noalign{}
\endhead
\bottomrule\noalign{}
\endlastfoot
Corpus \(A\) & \(a\) & \(c\) & \(n_A = a + c\) \\
Corpus \(B\) & \(b\) & \(d\) & \(n_B = b + d\) \\
Total & \(n_{word} = a + b\) & \(n_{\neg word} = c + d\) &
\(n = n_A + n_B\) \\
\end{longtable}

\emph{Log-Likelihood-Ratio} (\(LLR\)) measures deviation of the table
from what would be expected if the relative frequencies of \emph{word}
in \(Pop_A\) and \(Pop_B\) are the same.

\[LLR =  -2 \left( a \log\left(\frac{a}{E_a}\right) + b \log\left(\frac{b}{E_b}\right) +
c \log\left(\frac{c}{E_c}\right) + d \log\left(\frac{d}{E_d}\right) \right), \]

where \(E_a = \frac{a}{a+b} \cdot \frac{a}{a+c}\) and similiar for
\(E_b\), \(E_c\), \(E_d\).

Formally, \(LLR\) is a \emph{test statistic} for testing

\(H_0\): \(\pi_A = \pi_B\) vs.~\(H_1\): \(\pi_A \neq \pi_B\)

where \(\pi_A\) and \(\pi_B\) are the ``true'' frequencies, i.e.~the
frequencies in \(Pop_A\) and \(Pop_B\). \(\pi_A\) hence is the
probability that a single token drawn from \(Pop_A\) is equal to the
word under consideration and \((1-\pi_A)\) is the probability that it is
some other word. If \(H_0\) is true, i.e.~the probabilities (or
population frequencies) are the same, the distribution of \(LLR\) is
(approximately) known \emph{under the assumption of independence},
i.e.~under the model where corpus \(A\) is obtained by randomly drawing
tokens from \(Pop_A\), one by one, independent of each other, and
similarly for corpus \(B\). \(LLR\) then approximately follows a
\(\chi^2\)-distribution with 1 degree of freedom (\(\chi^2_1\)). Since
this distribution is known, it is easy to judge whether the
\(LLR\)-value calculated for a given word is within the range of what
one would expect by random variation or whether it is much larger.
Usually, cut-offs are chosen from quantiles of this distribution, for
example \(\chi^2_{1, 0.95}\) = 3.8414588, i.e.~if \(\pi_A = \pi_B\), the
\(LLR\)-score will only be larger than 3.84 with probability 5\% and in
this case one would declare the observed difference significant at the
5\% level. This means than whenever \(\pi_A = \pi_B\) a significant
difference is only declared 5\% of the time.

Equivalently, the p-value can be calculated as the probability for a
\(LLR\)-value as large or larger than the value that was actually
observed given that \(H_0\) is true, i.e.~\(\pi_A = \pi_B\). The result
is declared significant at level \(\alpha\) if the p-value is smaller
than or equal to \(\alpha\). A common value for \(\alpha\) is 0.05,
although \(\alpha\) should be much smaller when more than one word is
considered (as is the case in keyness analysis) and adjustments are
available in most software packages.

Even under the token-by-token sampling assumption, the
\(\chi^2_1\)-distribution of \(LLR\) under the null is only approximate.
The \(\chi^2\)-test for contingency tables uses a test statistic with a
different formula which however often results in similar values and also
approximately follows a \(\chi^2_1\)-distribution under the null. It has
been argued that the \(LLR\)-statistic is more appropriate for keyness
analysis than the \(\chi^2\)-test because the approximation is more
accurate in the case of skewness (Dunning 1993). In any case the model
assumes that the corpora are drawn independently token-by-token, which
in unrealistic, making the argument somewhat irrelevant.

While the (approximate) \(\chi^2_1\)-distribution can be derived
analytically, the true distribution can be approximated to arbitrary
precision using a \emph{permutation approach}, and this approach is
easily adapted for the more realistic assumption that tokens are not
drawn one-by-one but arrive in larger batches (one text at a time).

First we note that under the model where the tokens are drawn
independently one-by-one from \(Pop_A\) and \(Pop_B\), the row totals
\(n_A\) and \(n_B\) (corpus sizes) and the column totals \(n_{word}\)
and \(n_{\neg word}\) (total frequencies of the word under consideration
and all other words combined) do not contain \emph{any} information on
whether \(\pi_A = \pi_B\) as long as none of \(a\), \(b\), \(c\), \(d\)
is known. Hence, we can regard these as \emph{fixed}, although under the
model the total number of occurrences of a word across both corpora
would be subject to random fluctuations.

The distribution of \(LLR\) under \(H_0\), i.e.~assuming
\(\pi_A = \pi_B\) could now be obtained by performing (or simulating)
the following experiment:

\begin{enumerate}
\def\labelenumi{\arabic{enumi}.}
\tightlist
\item
  For every token in corpus \(A\) or \(B\) fill out a little sheet of
  paper. Put an ``X'' on the paper whenever the token is the word under
  consideration and leave the paper unmarked for any other word. This
  results in \(n\) slips of paper, \(n_{word}\) of these correspond to
  occurences of the word and \(n_{\neg word} = n - n_{word}\) to
  occurences of other words.
\item
  Put the \(n\) sheets into a big box and shuffle well.
\item
  Randomly draw \(n_A\) sheets from the box and note how many of these
  are marked. Put this number in the \(a\)-field of the contingency
  table.
\item
  Fill out the other fields of the table -- these can be obtained from
  \(a\) and the row and column totals.
\item
  Calculate the \(LLR\) score and record the value.
\item
  Repeat steps 2--5 a large number of times.
\end{enumerate}

The resulting empirical distribution is an approximation of the
theoretical distribution of \(LLR\) under \(H_0\). Alternatively, the
distribution can also be obtained numerically as given the row and
column totals, \(LLR\) is a function of \(a\) and under this model,
\(a\) follows a hypergeometrical distribution. The simulation approach,
however, can easily be adapted to the more realistic sampling model
described below.

Generally, sampling tokens one-by-one from two different population is a
poor model of how corpora are actually created. Usually, whole documents
are added to the corpus, i.e.~tokens arrive in larger sets. Also, many
words are distributed quite unevenly across the corpus and they often
occur in a few documents with a much higher frequency and not all in
others. Hence, adding a single document to a corpus leads to much
greater changes in a test statistic like \(LLR\) compared to
independently drawing the same number of tokens according to the
independence model described above.

We therefore propose the treat the corpora as samples of
\emph{documents}, not as samples of \emph{tokens} and assume random
sampling of documents, independently of each other. This is arguably
more realistic than the sampling-by-token model, although of course it
may also be an oversimplification in some situations. While the sampling
units are now documents, we still want to make statements about
frequencies of words in the populations (e.g.~in occurences per
million), not about frequencies per document.

We still want to test the hypothesis

\[
H_0: \pi_A = \pi_B \quad \text{vs.} \quad H_1: \pi_A \neq \pi_B    
\]

where \(\pi_A\) and \(\pi_B\) again are the frequencies of a given word
in \(Pop_A\) and \(Pop_B\), but we actually test the somewhat stronger
hypothesis

\[
H_0: (n_A, N_A-n_A) \stackrel{d}{=} (n_B, N_B-N_b),
\] i.e.~the pair of random variables (number of occurrences of
\emph{word} in Text, number of other \emph{tokens} in text) has the same
distribution among both \(Pop_A\) and \(Pop_B\). Apart from the
frequencies being the same this implies that the number of tokens per
text is not systematically different. See chapter 3 of Good (2005) for a
more technical discussion on assumptions for permutation tests.

Hence, for what follows, we assume that:

\begin{itemize}
\tightlist
\item
  Documents in both \(A\) and \(B\) are \emph{homogenous}, i.e.~not a
  mixture of different types
\item
  Documents in \(A\) and \(B\) are \emph{comparable} in size and type,
  e.g.~texts in \(A\) are not systematically longer than those in \(B\).
\item
  We have frequency information \emph{by document} for both corpora
  (e.g.~as a term-document-matrix)
\end{itemize}

We do \emph{not} need to assume that

\begin{itemize}
\tightlist
\item
  All the texts have the same length
\item
  \(A\) and \(B\) consist of the same number of texts
\item
  The order of tokens \emph{in} a text is in any way random
\end{itemize}

Examples where we would use the method include:

\begin{itemize}
\tightlist
\item
  \(A\) and \(B\) are corpora of texts from the same newspaper but cover
  different years
\item
  \(A\) and \(B\) are corpora of records of parliamentary debates from
  the same country but cover different years
\item
  \(A\) and \(B\) are corpora of two quality newspapers from the same
  year.
\item
  \(A\) and \(B\) are corpora of contemporary poetry written by men and
  women, respectively
\end{itemize}

Examples where it should probably not be used include:

\begin{itemize}
\tightlist
\item
  \(A\) and \(B\) are corpora of a tabloid and quality newspaper from
  the same year (debatable).
\item
  \(A\) is a corpus of poems and \(B\) a corpus of short stories from
  the same author.
\item
  \(A\) consists of written texts and \(B\) of transcriptions of spoken
  language.
\item
  \(A\) and \(B\) both consist of crawled web forums, tweets and
  newspaper articles from the same sources but different years.
\end{itemize}

Most of these cases could be treated with similar methods using more
sophisticated resampling schemes. These are part of ongoing research and
here, we will only focus on the simple case of two homogenuous corpora
which are similar in all other respects except for word frequencies.

The difference in the sampling scheme is now that we assume we sample
document-by-document, not token-by-token, i.e.~a whole set of tokens
arrives at a time and the number of occurrences of the word under
consideration is allowed to be much larger or much smaller compared to
what would be expected from randomly drawing token-by-token. After
sampling the two corpora document-by-document, we can create the same 2
x 2 - table as above by counting occurences and totals, and we can
calculate the same score (for example \(LLR\)) as in the token-by-token
case; the result will be the same.

The difference is the \emph{distribution} of the score, as the larger
fluctuation results in a wider distribution. This means that under this
model, scores that are regarded \emph{extreme} in the
token-by-token-sampling-model may well occur quite frequently randomly,
even if the population frequencies do not differ. It is not obvious how
the distribution of the score under the document-by-document-sampling
could be treated analytically, but it is possible to use a simulation
approach similar to the one described above for token-by-token sampling.

If \(\pi_A = \pi_B\), i.e.~there is no systematic difference in
frequencies in the populations, and under the assumption that the
document length do not differ systematically, a document with a given
total number of tokens and a given number of occurences of the word
under consideration could equally well come from \(Pop_A\) or \(Pop_B\).
Hence, the actually observed table or the score calculated from it
should not be very different from what would have been obtained when the
labels of the documents had been randomly assigned. We can then
approximate the distribution of \(LLR\) (or other statistics) by not
\emph{shuffling and randomly drawing tokens}, but by \emph{shuffling and
randomly drawing documents}. In contrast to the token-by-token model the
number of tokens in the randomly generated corpora is not fixed, but the
number of documents in the corpora is.

This results in the following experiment, which can easily be simulated
on a computer:

\begin{enumerate}
\def\labelenumi{\arabic{enumi}.}
\tightlist
\item
  For every \emph{document} in corpus \(A\) or \(B\) fill out a little
  sheet of paper. Put the number of occurences of the word under
  consideration and the total number of tokens in the document on the
  paper. This results in \(N = N_A + N_B\) slips of paper, where \(N_A\)
  and \(N_B\) are the numbers of documents in \(A\) and \(B\).
\item
  Put the \(N\) sheets into a big box and shuffle well.
\item
  Randomly draw \(N_A\) sheets from the box and add up the occurrences
  of the word under consideration as field \(a\) and the total number of
  tokens \(n_A\) in the table.
\item
  Fill out the other fields of the table.
\item
  Calculate the \(LLR\) score and record the value.
\item
  Repeat steps 2--5 a large number of times.
\end{enumerate}

We now use the empirical distribution of the \(LLR\) scores obtained in
this way as an approximation of the (unknown) theoretical distribution
of the scores. The extremeness of a given score value is now judged
against this distribution, which is typically considerably wider than
the \(\chi^2_1\) distribution or the simulated distribution obtained
under the token-by-token model. This is especially pronounced if the
occurrences are concentrated in a small number of documents.

\hypertarget{numerical-expriment-dail-corpus}{%
\section{Numerical Expriment: Dail
corpus}\label{numerical-expriment-dail-corpus}}

A numerical example shows that the actually much wider null distribution
is a considerable problem. We use a recently released corpus of
transcripts of parliamentary debates in Ireland (Herzog and Mikhaylov
2017). We assign the 919 transcripts from 2001 -- 2010 to Corpus \(A\)
and the 925 transcipts from 1991 -- 2000 to corpus \(B\). Since
different parliamentary sessions treat different topics, the different
dispersion of words between texts should be especially pronounced in
this example: many words should appear very frequently in some
transcripts and not at all in others.

We run the standard approach of calculating \(LLR\) scores for every
word. All in all, there are 193170 different words in both corpora
combined. About 40\% of these are significantly more frequent in \(A\)
or \(B\) (\(p < 0.05\)), 29\% have \(p < 0.01\), according to the
standard \(LLR\) approach (token-by-token) using a comparison with
quantiles of the \(\chi^2_1\) distribution.\\
We now \emph{shuffle the labels}: Of all 1844 texts, we randomly label
919 as \(A\) and the remaining 925 as \(B\). Now there are no
\emph{systematic} differences in word usage in \(A\) and \(B\), and
words can only appear more frequently in one of the corpora by chance
(so there are no true keywords here!).

We re-run the Log-Likelihood-Ratio test on these new randomized data. We
repeat this 100 times. This results in the following procedure:

\begin{enumerate}
\def\labelenumi{\arabic{enumi}.}
\tightlist
\item
  Calculate the test statistic (\(LLR\)) for every potential keyword
\item
  Shuffle the labels: Randomly assign the texts to \(A\) or \(B\) so
  that the same number is assigned to each corpus as in the original
  labelling
\item
  Calculate the test statistic (\(LLR\)) for every potential keyword
  based on the shuffled labels and record it
\item
  Repeat steps 2-3 a large number of times
\item
  For each potential keyword, obtain a p-value by comparing with the
  \(\chi^2_1\) distribution.
\end{enumerate}

We do expect a few \emph{false positives}: On average 5\% of the words
should have a p-value smaller than or equal to 0.05, 1\% should have a
p-value smaller than or equal to 0.01 etc. But, as Figure 1 shows, the
numbers of false positives are actually much larger when naively
comparing the values of \(LLR\) to a \(\chi^2_1\)-distribution (red)
vs.~comparing to the simulated null distribution based on
text-by-text-sampling (blue) as well as a permutation test based on the
log-ratio (see also section 4). We see that for most random
relabellings, the naive approach yields many more false positives than
expected, while for the permutation approaches, most random relabellings
do not produce a high number of false positives.

\begin{figure}
\centering
\includegraphics{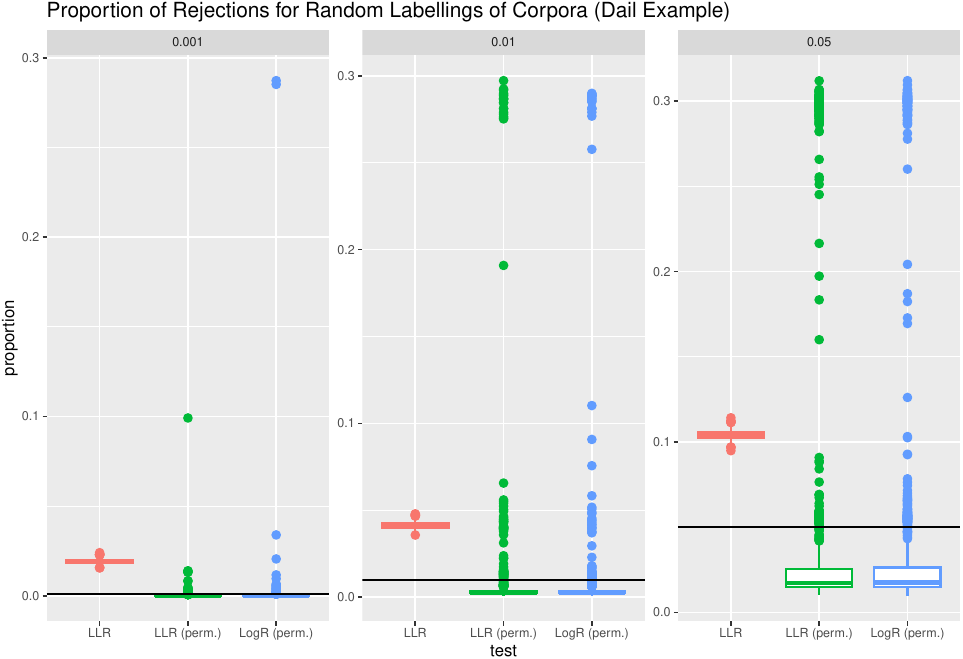}
\caption{Simulation results: Boxplots of the proportion of significant
words at different significance levels from randomly reassignment of the
texts to corpora \(A\) and \(B\) (dots) when the usual LLR-approach is
used (red) vs.~our permutation-based approach using LLR (blue) or the
log-ratio (green, see sec.~4).}
\end{figure}

We now look at the simulated null distribution for one word, ``simon''.
As a proper noun, we can expect this word to be quite unevenly
distributed, because a person named ``Simon'' could be mentioned
frequently in the same document and not at all in others. Figure 2 shows
the histogram of the simulated null distribution compared to the usually
used \(\chi^2_1\) distribution. We see that the simulated distribution
is wider, meaning that it frequently produces values that would be
considered extreme relative to the \(\chi^2_1\) distribution.

\begin{figure}
\centering
\includegraphics{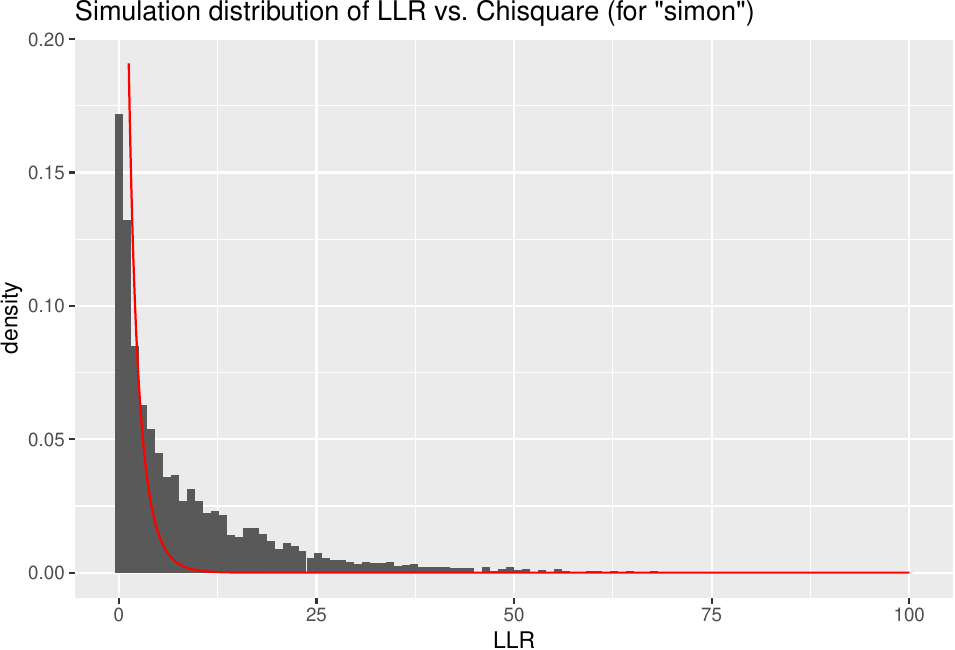}
\caption{Simulated null distribution (histogram) vs.~\(\chi^2_1\)
distribution (red)}
\end{figure}

Generally, the distribution of a test statistic under \(H_0\) can be
very different even if two words occur with the same frequencies in
\(A\) and \(B\). Very often, the distributions put more mass on larger
values than the \(\chi^2_1\). Especially when occurrences of a word are
concentrated in only a few texts (lumping), large values of \(LLR\) have
a much higher probability than they would according to the \(\chi^2_1\)
distribution. With our approach, the \emph{same} \(LLR\) value can lead
to \emph{very different} p-values for different words, and this is
desired!

Figure 3 shows p-values versus value of the \(LLR\) statistic for a
random selection of words from the corpora. The red curves are based on
the commonly used \(\chi^2_1\) distribution and are the same for all
words, while the blue ones ares based on the permutation approach. The
dots mark the realized value, and we see that p-values based on the
permutation approach are usually larger, sometimes much larger, than
those based on the \(\chi^2_1\)-distribution, meaning that under the
more realistic text-by-text sampling model, the
\(\chi^2_1\)-distribution (based on the token-by-token sampling model)
produces too many significant results.

\includegraphics{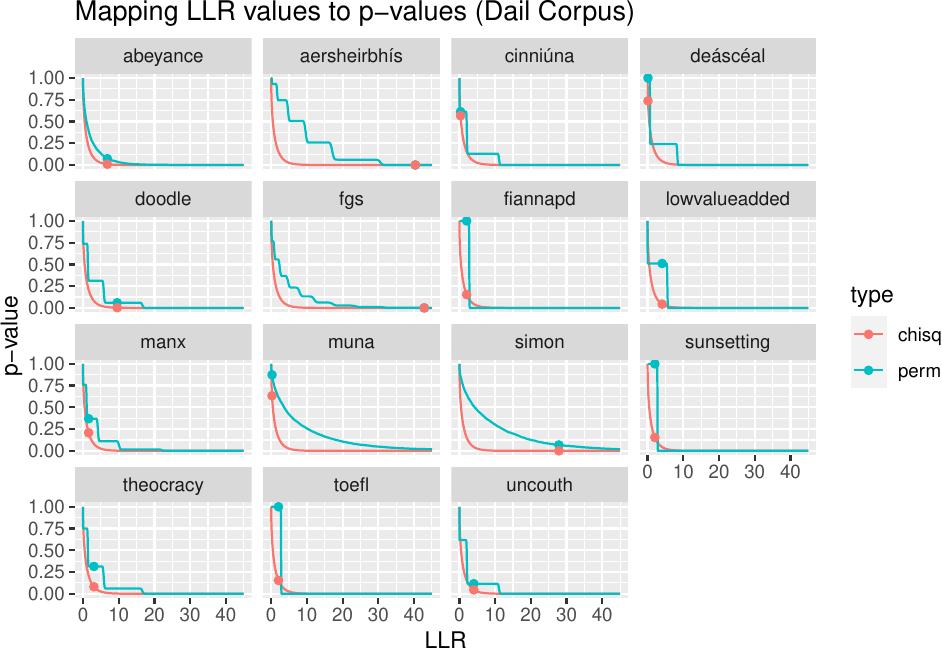}

Our approach is computationally costly. The test produces valid p-values
if the observed test statistic on the original data set is included in
the permutation distribution, but the test is a bit conservative in this
case and the minimum possible p-value is \(1/(n_{simulations} + 1)\).
This means that if a multiple testing correction is used, we are
especially interested in very small p-values and we need a huge number
of simulations to obtain a reasonable accuracy. This is considered in
more detail in the next section.

\hypertarget{implementation-and-extensions}{%
\section{Implementation and
extensions}\label{implementation-and-extensions}}

\hypertarget{choice-of-test-statistic}{%
\subsection{Choice of Test Statistic}\label{choice-of-test-statistic}}

The approach based on permutation tests allows us to approximate the
null distribution of \emph{any} test statistic related to keyness. This
includes the commonly used \(LLR\) and \(Chisquare\) statistics.
However, these are basically only used because their distribution under
the null hypothesis are approximately known -- under the inappropriate
token-by-token-sampling model. Otherwise, they have a number of
drawbacks:

\begin{itemize}
\tightlist
\item
  \(LLR\) and \(Chisquare\) are not very well interpretable, they are
  essentially measuring the deviation from a null model. In addition,
  this null model is itself also based on the inappropriate
  token-by-token-sampling assumption, making the interpretation of the
  measure questionable under more realistic assumptions.
\item
  \(LLR\) and \(Chisquare\) do not give an indication of the effect
  size, although \(Chisquare\)-based effect sizes are available in the
  literature.
\item
  \(LLR\) and \(Chisquare\) are directionless in the sense that they do
  not discriminate between \(\pi_A > \pi_B\) and \(\pi_A < \pi_B\). If
  only departures in one direction are to be detected, one has to resort
  to ad-hoc filtering based on observed frequencies.
\item
  \(LLR\) and \(Chisquare\) are based on 2x2 contingency tables and are
  symmetric with respect to occurrences and non-occurrences. In
  classical keyness analysis counting non-occurences is straightforward.
  If one is for example interested in constructs consiting of several
  words, the definition of non-occurences becomes difficult.
\end{itemize}

Since we need to simulate the distribution anyway, there is no reason
for using these test statistics. We can as well use a test statistic
that is more interpretable. An obvious candidate is the \emph{logratio}
statistic:

\[
Logratio = \log_2 \frac{\frac{a}{n_A}}{\frac{b}{n_B}}
\]

This statistic is well-established for keyness-analysis. It has a few
advantages over \(LLR\)/\(Chisquare\):

\begin{itemize}
\tightlist
\item
  It directly measures the effect size, and hence is very interpretable.
  An increase by one unit means that the ratio of the relative
  frequencies doubles.
\item
  The logratio gives the direction of the effect, positive values mean
  that the word is more frequent in \(A\), negative values mean it is
  more frequent in \(B\)
\item
  The logratio is based on the number of occurrences and the sizes of
  the corpora as measured by the number of tokens, hence it does not
  require the calculation of non-occurrences.
\end{itemize}

So far, the logratio has only been used as an effect size, as the
sampling distribution is not known, so it is commonly not used to assess
statistical significance, and indeed values can be large for very rare
words as it is easy to have a large increase in relative frequencies by
chance in this case. This is either taken care of by ad-hoc-filtering by
a minimum number of occurrences or by using some other
significance-based statistic like \(LLR\) for filtering. \texttt{CQPweb}
also offers the option to use an approximate confidence interval for the
Logratio for filtering. However, the calculation is also based on the
token-by-token sampling model.

For our approach, we can readily approximate the null distribution also
for the logratio statistic. It is important to note that the
distribution will also be different for each pair of corpora and for
each word under consideration, depending on the absolute number of
occurrences and the evenness of the distribution (dispersion). Two words
may well have the same logratio value, but the resulting p-values could
be very different, making one significant while the other is not.

We can then filter by significance using the p-values obtained using the
random permutations, and order the words surviving the filtering step by
the size of the logratio.

One problem with the logratio statistic is that it can take (positive or
negative) infinite values if one of the relative frequencies in the
numerator or the denominator is zero. Note that not both can be zero if
there is at least one occurrence of the word. This is unconvenient, as
zero occurences in Corpus \(B\) lead to a value of \(+\infty\),
regardless of whether the word appears in \(A\) exactly one time or
several thousand times. Even if the observed Logratio is finite,
infinite values can also easily occur when calculating the null
distribution, as occasionally the random permutation may assign all
documents containing the word to one of the corpora.

For this reason, it may be helpful to slightly modify the definition
using and add a typically small number \(k>0\):

\[
Logratio = \log_2 \frac{\frac{a+k}{n_A+k}}{\frac{b+k}{n_B+k}}
\] This amounts to adding \(k\) occurences of the word to both corpora
and also increasing the total number of tokens in both by \(k\);
\(k = 0\) corresponds to the original definition. The simplest choice is
\(k=1\), but \(k\) need not be an integer. This makes the logratio take
a finite value also when either \(a\) or \(b\) is zero. In the case
where \(b = 0\), the logratio will increase with \(a\), which
intuitively makes sense - a high keyness score should be assigned to a
word that occurs very often in \(A\) but not at all in \(B\), while one
occurrence in \(A\) and none in \(B\) is not an indicator of keyness.
The trick of adding a small amount to avoid zeros is often called a
\emph{Laplace-correction} and well known in statistics. The \emph{simple
math} statistic introduced in Kilgariff (2009) is -- apart from not
taking logarithms -- based on a similar idea, although it is only
suggested for use as a descriptive measure.

\hypertarget{implementation}{%
\subsection{Implementation}\label{implementation}}

Whether \(LLR\), \(Chisquare\), Logratio or any other statistics is
used, the basic steps to obtain a p-value are the following:

\begin{enumerate}
\def\labelenumi{\arabic{enumi}.}
\tightlist
\item
  Calculate the test statistic for every potential keyword
\item
  Shuffle the labels: Randomly assign the texts to \(A\) or \(B\) so
  that the same number is assigned to each corpus as in the original
  labeling
\item
  Calculate the test statistic for every potential keyword based on the
  shuffled labels and record it
\item
  Repeat steps 2-3 a large number of times
\item
  For each potential keyword, obtain a \emph{p-value} by calculating the
  fraction of values of the test statistic that are as extreme or more
  extreme than the one obtained for the original labeling.
\end{enumerate}

The implementation is conceptually most straightforward if the
frequencies are given as a term-document matrix \(T\), with counts for
term \(i\) in document \(j\) stored in \(T_{ij}\), although this is not
the most efficient data structure in terms of speed and memory usage. If
the columns are originally arranged such that columns \(1, \dots, N_A\)
correspond to the documents in \(A\) and columns
\(N_A+1, \dots, N_A+N_B\) correspond to those in \(B\), the shuffling of
documents corresponds to randomly permuting the columns and assigning
the first \(N_A\) columns to corpus \(A\) and the others to \(B\) before
re-calculating the statistic.

Note that we shuffle the documents (or columns of the term-document
matrix) once before re-calculating the statistics for \emph{all} terms
under consideration. In this way, also dependencies between occurrences
of different words are kept intact during resampling. While this is not
needed for the calculation of p-values, it is computationally more
efficient, and it may be of interest for other analyses. In addition,
some corrections for multiple testing require knowledge of correlations
between different p-values.

For the calculation of p-values we follow Chihara and Hesterberg (2019,
ch.~3.3): valid (slightly conservative) p-values are calculated easily
in the following way. For a one-sided test (right side), we count the
number of random permuations that resulted in a value of the test
statistic that was greater or equal to the observed value of the
original, unpermuted data: \[
\text{p-value}_{right} = \frac{\text{no. greater + no. equal + 1}}{\text{no. of permutations + 1}}
\] and similar for the left-sided test: \[
\text{p-value}_{left} = \frac{\text{no. less + no. equal + 1}}{\text{no. of permutations + 1}}
\] Adding \(1\) in both the numerator and denominator amounts to
including the observed value. This results in a slightly conservative
p-value, but guarantees that the test is valid for any number of random
permutations. It also means that never a p-value of zero is returned but
the minimum possible p-value is \(1/(\text{no. permutations} + 1)\).

The two-sided p-value is calculated by

\[
\text{p-value}_{twosided} = 2 \cdot \min \{\text{p-value}_{left}, \text{p-value}_{right} \}
\] (values larger than 1 are set to 1).

The approximation of the p-values by randomly drawing permutations is
more accurate if the number of iterations is larger. The construction as
given above (add 1 in numerator and denominator), however, ensures that
we err on the conservative side. If the null hypothesis is true, the
probability of obtaining a p-value smaller that \(\alpha\) is never
greater than \(\alpha\), although it may be considerably smaller than
\(\alpha\) if the number of random permutations is small. Hence, the
test is valid for any number of permutations, but the power may be low
if the number of permutations is small.

Since keyness analysis typically involves testing a large number of
words, some kind of multiple testing correction should be employed. For
this, p-values are compared to a much smaller threshold than the
conventionally used \(\alpha = 0.05\) for single tests. This means that
we need very high accuracy especially for the very small p-values, an in
addition, the minimum p-value than can possibly be obtained with a given
number of random draws is \(1/(\text{no. permutations} + 1)\) for
one-sided tests and \(2/(\text{no. permutations} + 1)\) in the two-sided
case. So the number of permutations should be chosen as large as
feasible (in the millions), but must in any case be large enough that it
is possible to obtain p-values smaller than the significance threshold.

This makes the method somewhat computationally costly, and an efficient
implementation is needed. The \texttt{keyperm} package for \texttt{R}
uses code partly written in \texttt{C++} that utilizes an efficient data
structure. This is made possible by use of the \texttt{Rcpp} package
(Eddebuettel et al.~2023, Eddelbuettel 2012). In addition, the
calculations can be trivially parallelized and results of several runs
on different cores can be easily combined. It should also be noted that
only the small p-values are needed with high accuracy; p-values far away
from any reasonable significance threshold need to be known only very
approximately. This suggests performing an initial run of only a few
thousand random permutations to decide on the words for which more
accurate p-values are needed. Only for these more extensive runs are
needed. The package also includes some helper functions and example code
to enable this.

\hypertarget{r-example}{%
\subsection{R Example}\label{r-example}}

We now give a simple example of use of the \texttt{keyperm} package
using small Reuters corpora. We first load the package, as well as the
\texttt{tm} package (Fleinerer and Hornik 2023) which includes the
example corpora. These are loaded as well:

\begin{Shaded}
\begin{Highlighting}[]
\FunctionTok{library}\NormalTok{(keyperm)}
\FunctionTok{library}\NormalTok{(tm)}

\CommentTok{\# load subcorpora "acq" and "crude" from Reuters}

\FunctionTok{data}\NormalTok{(acq)}
\FunctionTok{data}\NormalTok{(crude)}
\end{Highlighting}
\end{Shaded}

We then calculate a term-document matrix for both corpora separately and
combine them into one \texttt{tdm} object. Currently, \texttt{tdm}
objects using the \texttt{tm} package are the only supported input
format for the \texttt{keyperm} package. We then also create a logical
vector that indicates which columns of the term-document matrix belong
to which corpus.

\begin{Shaded}
\begin{Highlighting}[]
\CommentTok{\# convert to term{-}document{-}matrices and combine into single tdm}

\NormalTok{acq\_tdm }\OtherTok{\textless{}{-}} \FunctionTok{TermDocumentMatrix}\NormalTok{(acq, }\AttributeTok{control =} \FunctionTok{list}\NormalTok{(}\AttributeTok{removePunctuation =} \ConstantTok{TRUE}\NormalTok{))}
\NormalTok{crude\_tdm }\OtherTok{\textless{}{-}} \FunctionTok{TermDocumentMatrix}\NormalTok{(crude, }\AttributeTok{control =} \FunctionTok{list}\NormalTok{(}\AttributeTok{removePunctuation =} \ConstantTok{TRUE}\NormalTok{))}
\NormalTok{tdm }\OtherTok{\textless{}{-}} \FunctionTok{c}\NormalTok{(acq\_tdm, crude\_tdm)}

\CommentTok{\# generate a logical that indicates whether document comes from "acq" or "crude"}

\NormalTok{ndoc\_A }\OtherTok{\textless{}{-}} \FunctionTok{dim}\NormalTok{(acq\_tdm)[}\DecValTok{2}\NormalTok{]}
\NormalTok{ndoc\_B }\OtherTok{\textless{}{-}} \FunctionTok{dim}\NormalTok{(crude\_tdm)[}\DecValTok{2}\NormalTok{]}
\NormalTok{corpus }\OtherTok{\textless{}{-}} \FunctionTok{rep}\NormalTok{(}\FunctionTok{c}\NormalTok{(}\ConstantTok{TRUE}\NormalTok{, }\ConstantTok{FALSE}\NormalTok{), }\FunctionTok{c}\NormalTok{(ndoc\_A, ndoc\_B))}
\end{Highlighting}
\end{Shaded}

Now we convert the \texttt{tdm} object to what we call an indexed
frequency list, containing the same information but in an optimized data
structure especially suitable for fast computations.

\begin{Shaded}
\begin{Highlighting}[]
\CommentTok{\# generate an indexed frequency list, the data structure used by keyperm}

\NormalTok{reuters\_ifl }\OtherTok{\textless{}{-}} \FunctionTok{create\_ifl}\NormalTok{(tdm, }\AttributeTok{corpus =}\NormalTok{ corpus)}
\end{Highlighting}
\end{Shaded}

We now calculate the \(LLR\) values along with p-values from the
conventionally used \(\chi^2_1\)-distribution, which -- as we argued
above -- is wrong because it is based on a token-by-token sampling
model.

\begin{Shaded}
\begin{Highlighting}[]
\CommentTok{\# calculate Log{-}Likelihood{-}Ratio scores for all terms and calculate}
\CommentTok{\# p{-}values according to the (wrong) token{-}by{-}token sampling model}

\NormalTok{llr }\OtherTok{\textless{}{-}} \FunctionTok{keyness\_scores}\NormalTok{(reuters\_ifl, }\AttributeTok{type =} \StringTok{"llr"}\NormalTok{, }\AttributeTok{laplace =} \DecValTok{0}\NormalTok{)}
\FunctionTok{head}\NormalTok{(}\FunctionTok{round}\NormalTok{(}\FunctionTok{pchisq}\NormalTok{(llr, }\AttributeTok{df =} \DecValTok{1}\NormalTok{, }\AttributeTok{lower.tail =} \ConstantTok{FALSE}\NormalTok{), }\AttributeTok{digits =} \DecValTok{4}\NormalTok{), }\AttributeTok{n =} \DecValTok{10}\NormalTok{)}
\end{Highlighting}
\end{Shaded}

\begin{verbatim}
##        125        150     200000      50000    acquire additional       also 
##     0.1072     0.3886     0.9483     0.3523     0.0003     0.1884     0.2315 
##        and        any        are 
##     0.1437     0.4504     0.3589
\end{verbatim}

Now we obtain permutation-based p-values using the \texttt{keyperm()}
function, which are usually, but not always, larger. We pass the indexed
frequency list, and the original \(LLR\) values and indicate that we
want 10000 permutation values of the \(LLR\) statistic:

\begin{Shaded}
\begin{Highlighting}[]
\CommentTok{\# generate permutation distribution and p{-}values based on document{-}by{-}document sampling model}

\NormalTok{keyp }\OtherTok{\textless{}{-}} \FunctionTok{keyperm}\NormalTok{(reuters\_ifl, llr, }\AttributeTok{type =} \StringTok{"llr"}\NormalTok{, }
                \AttributeTok{laplace =} \DecValTok{0}\NormalTok{, }\AttributeTok{output =} \StringTok{"counts"}\NormalTok{, }\AttributeTok{nperm =} \DecValTok{10000}\NormalTok{)}
\FunctionTok{head}\NormalTok{(}\FunctionTok{p\_value}\NormalTok{(keyp, }\AttributeTok{alternative =} \StringTok{"greater"}\NormalTok{), }\AttributeTok{n =} \DecValTok{10}\NormalTok{)}
\end{Highlighting}
\end{Shaded}

\begin{verbatim}
##        125        150     200000      50000    acquire additional       also 
## 0.05139486 0.69863014 0.95930407 0.34696530 0.00489951 0.34696530 0.38326167 
##        and        any        are 
## 0.17118288 0.52484752 0.40865913
\end{verbatim}

We can also get p-values not using the \(LLR\) but the log-ratio. To
avoid dividing by zero, we use a laplace correction adding \(1\) both in
the numerator as well as the denominator. We do a first run with 1000
permutations:

\begin{Shaded}
\begin{Highlighting}[]
\CommentTok{\# generate observed log{-}ratio values and (one{-}sided) p{-}values based}
\CommentTok{\# on the permutation distribution (document{-}by{-}document sampling model)}
\CommentTok{\# laplace{-}correction used (adding one occurence to both corpora)}

\NormalTok{logratio }\OtherTok{\textless{}{-}} \FunctionTok{keyness\_scores}\NormalTok{(reuters\_ifl, }\AttributeTok{type =} \StringTok{"logratio"}\NormalTok{, }\AttributeTok{laplace =} \DecValTok{1}\NormalTok{)}
\NormalTok{keyp2 }\OtherTok{\textless{}{-}} \FunctionTok{keyperm}\NormalTok{(reuters\_ifl, logratio, }\AttributeTok{type =} \StringTok{"logratio"}\NormalTok{, }
                \AttributeTok{laplace =} \DecValTok{1}\NormalTok{, }\AttributeTok{output =} \StringTok{"counts"}\NormalTok{, }\AttributeTok{nperm =} \DecValTok{1000}\NormalTok{)}
\FunctionTok{head}\NormalTok{(}\FunctionTok{p\_value}\NormalTok{(keyp2, }\AttributeTok{alternative =} \StringTok{"greater"}\NormalTok{), }\AttributeTok{n =} \DecValTok{10}\NormalTok{)}
\end{Highlighting}
\end{Shaded}

\begin{verbatim}
##         125         150      200000       50000     acquire  additional 
## 0.023976024 0.676323676 0.557442557 0.064935065 0.003996004 0.064935065 
##        also         and         any         are 
## 0.209790210 0.096903097 0.256743257 0.781218781
\end{verbatim}

We now filter the small p-values, and run 9000 further simulations for
the corresponding words, as we need higher accuracy in the small
p-values. Note that 10000 simulations are usually not enough for real
practical applications.

\begin{Shaded}
\begin{Highlighting}[]
\NormalTok{pvals }\OtherTok{\textless{}{-}} \FunctionTok{p\_value}\NormalTok{(keyp2, }\AttributeTok{alternative =} \StringTok{"greater"}\NormalTok{)}
\FunctionTok{table}\NormalTok{(pvals }\SpecialCharTok{\textgreater{}} \FloatTok{0.1}\NormalTok{)}
\end{Highlighting}
\end{Shaded}

\begin{verbatim}
## 
## FALSE  TRUE 
##  1330  1042
\end{verbatim}

\begin{Shaded}
\begin{Highlighting}[]
\NormalTok{small\_p }\OtherTok{\textless{}{-}} \FunctionTok{which}\NormalTok{(pvals }\SpecialCharTok{\textless{}} \FloatTok{0.1}\NormalTok{)}

\CommentTok{\# subset the original logratio values and create a new, smaller, indexed frequency list:}

\NormalTok{logratio\_subset }\OtherTok{\textless{}{-}}\NormalTok{ logratio[small\_p]}
\NormalTok{reuters\_ifl\_subset }\OtherTok{\textless{}{-}} \FunctionTok{create\_ifl}\NormalTok{(tdm, }\AttributeTok{subset\_terms =}\NormalTok{ small\_p, }\AttributeTok{corpus =}\NormalTok{ corpus)}

\NormalTok{keyp2\_subset }\OtherTok{\textless{}{-}} \FunctionTok{keyperm}\NormalTok{(reuters\_ifl\_subset, logratio\_subset, }\AttributeTok{type =} \StringTok{"logratio"}\NormalTok{, }
                 \AttributeTok{laplace =} \DecValTok{1}\NormalTok{, }\AttributeTok{output =} \StringTok{"counts"}\NormalTok{, }\AttributeTok{nperm =} \DecValTok{9000}\NormalTok{)}
\end{Highlighting}
\end{Shaded}

We can use the \texttt{combine\_results()} function to combine the
results from both simulation runs. Note that this works despite the fact
that in the second simulation only a subset of words was used. The
function is also useful for parallelization where different simulation
runs may run on different cores or computers.

\begin{Shaded}
\begin{Highlighting}[]
\CommentTok{\# combine counts from both runs using the combiner}

\NormalTok{keyp2\_combined }\OtherTok{\textless{}{-}} \FunctionTok{combine\_results}\NormalTok{(keyp2, keyp2\_subset)}

\CommentTok{\# smaller p{-}values are based on 1000, the larger ones on 10000 random permutations}
\CommentTok{\# note that 10000 is still far too small for real applications}

\FunctionTok{head}\NormalTok{(}\FunctionTok{p\_value}\NormalTok{(keyp2\_combined, }\AttributeTok{alternative =} \StringTok{"greater"}\NormalTok{), }\AttributeTok{n =} \DecValTok{10}\NormalTok{)}
\end{Highlighting}
\end{Shaded}

\begin{verbatim}
##        125        150     200000      50000    acquire additional       also 
## 0.01679832 0.67632368 0.55744256 0.05479452 0.00259974 0.05479452 0.20979021 
##        and        any        are 
## 0.09549045 0.25674326 0.78121878
\end{verbatim}

\hypertarget{discussion-and-outlook}{%
\section{Discussion and Outlook}\label{discussion-and-outlook}}

We presented a permutation test approach to keyness analysis, based on a
\emph{text-by-text} sampling model, in contrast to the conventionally
used methods, which are implicitly based on a \emph{token-by-token}
sampling model. Unevenness of distrbiution of words makes the more
realtistic \emph{text-by-text} sampling distribution of a test statistic
typically wider than the conventionally used reference distributions,
meaning that seemingly extreme values of test statistics are actually
much more common than predicted by e.g.~the \(\chi^2_1\) distribution.

The idea can easily be extendended, and indeed proposals based on
similar ideas have been put forward by Gries (2006, 2022). For example,
if the two different corpora are from different years and both contain
tweets and newspaper articles (a case we excluded in our discussion
above), we could shuffle lables between tweets and articles separately,
not mixing the two, hence keeping the number of tweets and articles
constant in each resampling step. Also possible would be the use of test
statistics which compare more than two corpora at once.

Computationally similar to permutation tests but conceptually different
are bootstrap methods, which could be implemented similarly (see Gries
2022). These may be used to construct confidence intervals of a measure
instead of performing a test, and they could also be used for one-sample
problems, as sometimes it may be appropriate to treat a reference corpus
as fixed and only the corpus compared with it as a random sample.

Currently, only the basic version of the permutation test is implemented
in our \texttt{R} package \texttt{keyperm}, but the extensions sketched
here are part of ongoing investigations and some of these may be added
at a later date. Also, the package is currently available on github but
submission to CRAN is planned in the near future.

\hypertarget{references}{%
\section*{References}\label{references}}
\addcontentsline{toc}{section}{References}

\noindent Chihara, L.M., Hesterberg, T. (2019): \emph{``Mathematical
Statistics with Resampling and R''}, 2nd ed., Wiley, Hoboken. Dunning,
T. (1993): ``Accurate Methods for the Statistics of Surprise and
Coincidence'', \emph{Computational Linguistics 19(1)}, 61-74.\\
Eddelbuettel, D., Francois, R., Allaire, J., Ushey, K., Kou, Q.,
Russell, N., Ucar, I., Bates, D., Chambers, J. (2023). \emph{Rcpp:
Seamless R and C++ Integration}. R package version 1.0.11,
\url{https://CRAN.R-project.org/package=Rcpp}.\\
Eddelbuettel, D. (2013). \emph{Seamless R and C++ Integration with
Rcpp}. Springer, New York. \url{doi:10.1007/978-1-4614-6868-4}
\url{https://doi.org/10.1007/978-1-4614-6868-4}, ISBN
978-1-4614-6867-7.\\
Evert, S. (2006): ``How Random is a Corpus: The Library Metaphor'',
\emph{Zeitschrift für Anglistik und Amerikanistik 54(2)}, 177-190.\\
Feinerer, I., Hornik, K. (2023): \emph{tm: Text Mining Package}. R
package version 0.7-11, \url{https://CRAN.R-project.org/package=tm}.\\
Good, P. (2005): \emph{``Permutation, Parametric, and Bootstrap Tests of
Hypotheses''}, 3rd ed., Springer, New York.\\
Gries, S.T. (2008): ``Dispersions and adjusted frequencies in corpora'',
\emph{International Journal of Corpus Linguistics} 13:4, 403--437\\
Gries, S.T. (2022): ``Toward more careful corpus statistics: uncertainty
estimates for frequencies, dispersions, association measures, and
more'', \emph{Research Methods in Applied Linguistics} 1, 100002\\
Herzog, A., Mikhaylov, S.J. (2017): ``Database of Parliamentary Speeches
in Ireland, 1919--2013'', arXiv:1708.04557.v1\\
Kilgariff, A. (2005): ``Language is never, ever, ever, random'',
\emph{Corpus Linguistics and Linguistic Theory} 1-2, 263-276\\
Kilgariff. A. (2009). ``Simple maths for keywords''. In:
\emph{Proceedings of Corpus Linguistics Conference CL2009}, Mahlberg,
M., González-Díaz, V. \& Smith, C. (eds.), University of Liverpool, UK,
July 2009.

\end{document}